\pdfoutput=1

\documentclass[11pt]{article}

\usepackage[preprint]{acl}

\usepackage{times}
\usepackage{latexsym}
\usepackage{float} 
\usepackage{verbatim}

\usepackage[T1]{fontenc}

\usepackage[utf8]{inputenc}

\usepackage{microtype}

\usepackage{inconsolata}

\usepackage{graphicx}
\usepackage{enumitem}
\usepackage{amsmath}

\title{Akan Cinematic Emotions (AkaCE): A Multimodal Multi-party Dataset\\for Emotion Recognition in Movie Dialogues}

\author{
 \textbf{David Sasu\textsuperscript{1}},
 \textbf{Zehui Wu\textsuperscript{2}},
 \textbf{Ziwei Gong\textsuperscript{2}},
 \textbf{Run Chen\textsuperscript{2}},
 \textbf{Pengyuan Shi\textsuperscript{2}},
 \\
 \textbf{Lin Ai\textsuperscript{2}},
 \textbf{Julia Hirschberg\textsuperscript{2}},
 \textbf{Natalie Schluter\textsuperscript{1,}\thanks{Currently at Apple.}}
\\
 \textsuperscript{1}IT University of Copenhagen,
 \textsuperscript{2}Columbia University
}

\begin{document}
\maketitle
\begin{abstract}

In this paper, we introduce the Akan Conversation Emotion (AkaCE) dataset, the first multimodal emotion dialogue dataset for an African language, addressing the significant lack of resources for low-resource languages in emotion recognition research. AkaCE, developed for the Akan language, contains 385 emotion-labeled dialogues and 6,162 utterances across audio, visual, and textual modalities, along with word-level prosodic prominence annotations. The presence of prosodic labels in this dataset also makes it the first prosodically annotated African language dataset.
We demonstrate the quality and utility of AkaCE through experiments using state-of-the-art emotion recognition methods, establishing solid baselines for future research. 
We hope AkaCE inspires further work on inclusive, linguistically and culturally diverse NLP resources.
\end{abstract}


\section{Introduction}
Emotion Recognition in Conversation (ERC) is a rapidly evolving subfield of Natural Language Processing (NLP) that focuses on detecting or classifying the emotional states expressed by speakers in multi-turn conversations \cite{poria2019emotion}. Unlike traditional emotion recognition tasks that aim to identify emotions from isolated text or speech snippets or speech utterances such as \cite{zahiri2018emotion}, ERC seeks to leverage contextual cues from prior dialogue, speaker relationships, and conversational flow to infer emotional states more accurately \cite{poria2019emotion}.

In recent years, ERC has garnered significant attention within the NLP community, driven by its growing relevance to a range of real-world applications. Notable examples include empathetic chatbot systems \cite{fragopanagos2005emotion}, call-center dialogue systems \cite{danieli2015emotion}, and mental health support tools \cite{ringeval2018avec}. These systems rely on ERC to capture the evolving emotional dynamics of conversations, enabling more contextually appropriate and emotionally aware responses. Developing robust ERC systems often requires multimodal data integration \cite{poria2018meld}, which is challenging due to the need to jointly model diverse inputs like scene context, discussion topics, conversational history, and speaker personalities \cite{10.1145/3394171.3413909, hazarika-etal-2018-icon, wu-etal-2024-multimodal}. However, comprehensive multimodal ERC dialogue datasets remain scarce, with benchmark resources like IEMOCAP \cite{busso2008iemocap}, MSP-IMPROV \cite{busso2016msp}, MELD \cite{poria2018meld}, and M³ED \cite{zhao2022m3ed} being notable exceptions.

\begin{figure}[t]
    \centering
    \includegraphics[width=0.48\textwidth]{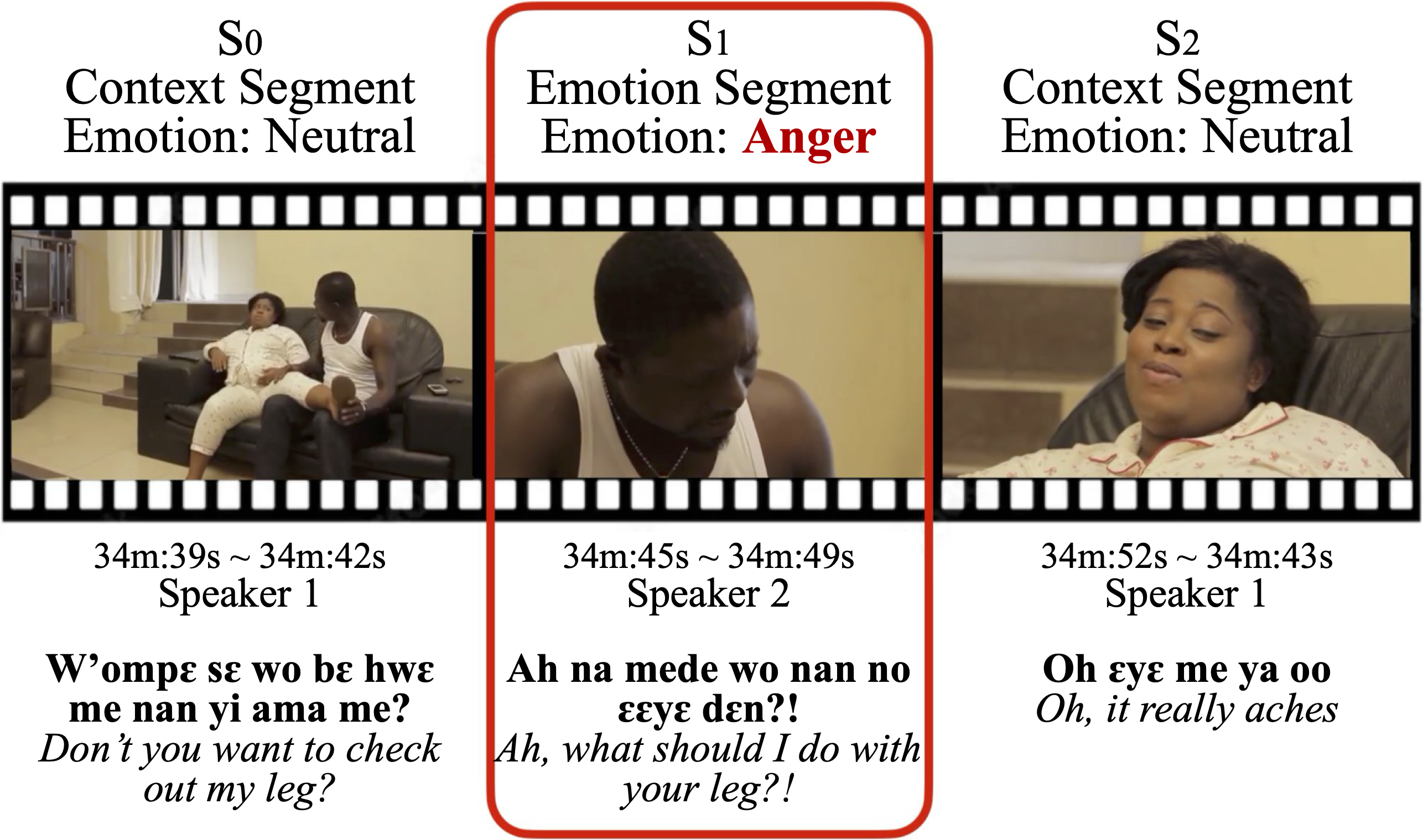}
    \caption{An example of a dialogue showing conversational context and emotion labels.}
    \label{fig:example}
\end{figure}

A major limitation of existing ERC datasets is their focus on high-resource languages, particularly English (IEMOCAP, MSP-IMPROV, MELD) and Chinese (M³ED). This lack of linguistic diversity hinders the development of ERC systems for low-resource languages, especially in Africa. To our knowledge, no multimodal ERC dataset exists for any African language, despite the continent being home to approximately 3,000 of the world’s 7,000 languages \cite{leben2018languages} and 18.3\% of the global population \cite{moibrahim2023facts}.

To address this gap, we introduce the Akan Conversation Emotion (AkaCE) dataset, a multimodal emotion dialogue dataset for Akan, a major West African language spoken by about 20 million people \cite{lctlresources}. Akan is the most widely spoken language in Ghana, with around 80\% of the population using it as a first or second language, and approximately 44\% identifying as native Akan speakers. It is also natively spoken in parts of Ivory Coast and Togo. The language primarily comprises three main dialects: Asante, Akuapem, and Fante.

AkaCE contains 385 emotion-labeled dialogues from 21 Akan movies, covering diverse scenes and topics. It includes 6,162 utterances from 308 speakers (155 male, 153 female), ensuring a gender-balanced dataset. As a tonal language, Akan’s prosodic features are crucial for emotion recognition, so AkaCE includes word-level prosodic prominence annotations to support research on prosody in ERC. Our baseline experiments validate the dataset’s quality and utility for low-resource and cross-cultural emotion recognition research. Our main contributions are:

\begin{itemize}[nosep,topsep=0pt]
    \item[1.] We introduce AkaCE, the first multimodal emotion dialogue dataset for an African language, enabling cross-cultural emotion research\footnote{Our data and code are available at \href{https://github.com/sasudavid/AkaCE/tree/main}{this GitHub repo}.}.
    \item[2.] We validate AkaCE through experiments with state-of-the-art ERC methods, establishing a strong baseline and detailed analysis.
    \item[3.] We provide word-level prosodic prominence annotations, making AkaCE the first prosodically annotated dataset for an African language, facilitating research on prosody’s role in ERC and tonal language processing.
\end{itemize}


\section{Related work}

\subsection{Related Datasets}

ERC and Speech Emotion Recognition (SER) datasets are essential for developing models to understand and classify emotions in speech. These datasets can be grouped into three main categories. Table~\ref{tab:datasets} compares AkaCE with other discussed datasets.

\textbf{Single-modality datasets} focus exclusively on a single modality, such as text, including EmoryNLP \cite{zahiri2018emotion}, EmotionLines \cite{chen1802emotionlines}, and DailyDialog \cite{li2017dailydialog}. These are useful for text-based emotion recognition but lack the multi-modal richness needed for comprehensive analysis involving vocal or visual cues.

\textbf{Multi-modal datasets} combine text, audio, or video, such as CMU-MOSEI \cite{zadeh2018multimodal}, AFEW \cite{fromcollecting}, MEC \cite{li2018mec}, and CH-SIMS \cite{yu2020ch}. While offering richer features, they are not conversational and miss the dynamic, context-dependent expressions seen in natural dialogues.

\textbf{Conversational multi-modal datasets} integrate text, audio, and video with conversational context, such as IEMOCAP \cite{busso2008iemocap}, MSP-IMPROV \cite{busso2016msp}, MELD \cite{poria2018meld}, and M³ED \cite{zhao2022m3ed}. But these datasets mainly focus on high-resource languages, leaving gaps for low-resource languages.

Existing datasets lack resources for low-resource languages, such as Akan, and prosodic annotations critical for tonal languages. AkaCE fills this gap by offering a multi-modal resource with prosodic labels and conversational data, enabling robust SER for Akan and advancing cross-cultural SER research. 


\begin{table*}[t]
\centering
\resizebox{\textwidth}{!}{%
\begin{tabular}{lccccccccc}
\hline
\textbf{Dataset} & \textbf{Dialogue} & \textbf{Modalities} & \textbf{Prosodic Annotations} & \textbf{Sources} & \textbf{Mul-label} & \textbf{Emos} & \textbf{Spks} & \textbf{Language} & \textbf{Utts} \\ \hline
EmoryNLP \cite{zahiri2018emotion} & Yes & $t$ & No & Friends TV & Yes & 9 & -- & English & 12,606 \\
EmotionLines \cite{chen1802emotionlines} & Yes & $t$ &  No & Friends TV & No & 7 & -- & English & 29,245 \\
DailyDialog \cite{li2017dailydialog} & Yes & $t$ &  No & Daily & No & 7 & -- & English & 102,979 \\\hline
CMU-MOSEI \cite{zadeh2018multimodal} & No & $a, v, t$ &  No & YouTube & No & 7 & 1000 & English & 23,453 \\
AFEW \cite{fromcollecting} & No & $a, v$ &  No & Movies & No & 7 & 330 & English & 1,645 \\
MEC \cite{li2018mec}& No & $a, v$ &  No & Movies, TVs & No & 8 & -- & Mandarin & 7,030 \\
CH-SIMS \cite{yu2020ch} & No & $a, v, t$ &  No & Movies, TVs & No & 5 & 474 & Mandarin & 2,281 \\\hline
IEMOCAP \cite{busso2008iemocap} & Yes & $a, v, t$ &  No & Act & No & 5 & 10 & English & 7,433 \\
MSP-IMPROV \cite{busso2016msp} & Yes & $a, v, t$ &  No & Act & No & 5 & 12 & English & 8,438 \\
MELD \cite{poria2018meld} & Yes & $a, v, t$ &  No & Friends TV & No & 7 & 407 & English & 13,708 \\
M³ED \cite{zhao2022m3ed} & Yes & $a, v, t$ &  No & 56 TVs & Yes & 7 & 626 & Mandarin & 24,449 \\
\textbf{AkaCE (Ours)} & Yes & $a, v, t$ & Yes & 21 Movies & No & 7 & 308 & Akan & 6,162 \\\hline
\end{tabular}%
}
\caption{Comparison of existing benchmark datasets. $a, v, t$ refer to audio, visual, and text modalities respectively.}
\label{tab:datasets}
\end{table*}

\subsection{Related Methods}

Conversational emotion recognition (ERC) has evolved through various approaches addressing contextual modeling, multimodal integration, and speaker dependencies. Early works used hierarchical LSTMs \cite{poria2017context} and Conversational Memory Networks (CMN) \cite{hazarika2018conversational} to capture context and inter-speaker influences, improving sentiment classification but struggling with generalization and sparse contexts.

DialogueRNN \cite{majumder2019attentive} and HiGRU \cite{jiao2019higru} refined speaker-specific emotion tracking and attention-based modeling but faced challenges with subtle distinctions and multimodal integration. Knowledge-enriched models \cite{zhong2019knowledge} leveraged commonsense knowledge for emotion detection but struggled with closely related emotions and low-resource settings.

Graph-based methods such as ConGCN \cite{zhang2019modeling} and DialogueGCN \cite{ghosal2019dialoguegcn} modeled multi-speaker dependencies effectively but relied heavily on textual features. Multimodal transformers like MulT \cite{tsai2019multimodal} and MMGCN \cite{hu2021mmgcn} advanced cross-modal fusion but faced scalability issues due to dataset alignment and computational demands.

Recent transformer-based models like DialogXL \cite{shen2021dialogxl} and EmoBERTa \cite{kim2021emoberta} improved ERC with dialog-aware attention and speaker-aware features but lacked multimodal capabilities. M2FNet \cite{chudasama2022m2fnet} addressed multimodal fusion, effectively integrating text, audio, and visual data, though it struggled with imbalanced datasets. Recent methods leverage LLMs to enhance performance, reformulating emotion recognition as a generative task \cite{lei2023instructerc}, incorporating acoustic features \cite{wu2024beyond} and contextual information \cite{xue2024bioserc, fu2024ckerc, zhang2023dialoguellm}.

Despite these advancements, existing methods often lack robust solutions for underrepresented languages and datasets. Our work bridges these gaps by introducing a multimodal dataset and a focus on low-resource settings, enabling more comprehensive and inclusive ERC research.

\section{AkaCE Dataset}
We construct the AkaCE dataset by collecting and annotating dialogues from Akan-language movies, with examples illustrated in Figure \ref{fig:example}. The dataset includes transcriptions with speaker identifications, emotion labels, and word-level prosodic prominence annotations. Table \ref{tab:datasets} compares AkaCE with other discussed datasets.

\subsection{Data Selection}
The dataset consists of 21 Akan movies that were downloaded from the Internet Archive. To ensure that the movies included within this dataset were of high quality we ensured that each of the movies selected to be a part of the dataset fulfilled the following criteria: (1) the movie must be complete and not truncated in any section, (2) the speech of the actors within the movie should be intelligible, (3) the facial expressions of the actors within movie should be clear. 

\subsection{Annotators and Annotation Process}

The annotation task was carried out by Akan data annotation professionals contracted through an institute of linguistics and bible translation in Ghana. The annotators consisted of five men and two women, all native Akan speakers. Of these seven annotators, three were employed to work full-time while the rest worked part-time. One of the full-time annotators opted to annotate seven movies, whereas the other two full-time annotators each chose to annotate five movies. The remaining four part-time annotators annotated one movie each. The movies were randomly assigned to their respective annotators. 

The data annotators recorded the desired data by watching the movies and simultaneously recording the necessary information into Microsoft Excel sheets. All resulting sheets were then collated into one Excel sheet, where all redundant entries were eliminated and annotation errors were corrected.

\subsection{Text and Speaker Annotation}
Even though there have been recent advances in Akan Automatic Speech Recognition (ASR), most modern Akan ASR systems still generate many recognition errors as a result of the dearth of training data available \cite{dossou2024advancingafricanaccentedspeechrecognition}. As a result of this, all the speech utterances for each movie were manually transcribed by the annotators before any emotion labelling was performed. 


Additionally, due to the lack of acoustic models for Akan that could facilitate audio alignment and automatically generate timestamps for each utterance in a movie, annotators manually tracked and recorded the timestamps for all utterances.

For the speaker annotations, the speaker for each utterance was identified by a unique identifier which consists of a combination of the order in which the speaker first appeared in the current dialogue and their gender. For instance, a possible label that would be assigned to a man who is the first speaker in the current dialogue of a scene within a movie is \texttt{`speaker one man'}.   

To ensure the high quality of the utterance transcriptions, a professional Akan linguist from the same institute was employed to peruse all of the transcriptions provided by the annotators and correct any identified errors.

\subsection{Emotion Annotation}
The emotion demonstrated for each utterance within a dialogue was annotated using one of seven possible emotion labels: \texttt{Sadness, Fear, Anger, Surprise, Disgust, Happy} and \texttt{Neutral}. Six out of these emotions (i.e Sadness, Fear, Anger, Surprise, Happy and Disgust) were proposed by Paul Ekman (\citeyear{ekman1992there}) as the six universal human emotions. Following previous works \cite{poria2018meld,busso2008iemocap,gong-etal-2024-mapping}, a neutral emotion label was added to identify utterances within the dataset that did not carry any pronounced emotional undertone. This set of seven emotion labels is further supported by \citet{ahmad2025exploring}, who employed the same categories in their study of cultural variation in emotion perception across 15 African languages.

The annotators were instructed to assign emotion labels to each utterance while simultaneously viewing their assigned movies. To ensure the accuracy and reliability of annotations, a preliminary information session was held by a research coordinator at the aforementioned institute in Ghana. This session provided a comprehensive overview of the annotation task, clarified expectations, and included illustrative examples of how each target emotion might manifest in various scenarios. In cases of uncertainty, annotators were guided to select the emotion label they deemed most appropriate for the utterance. The emotion annotation tutorial was designed with inspiration and reference to established emotion annotation guidelines, such as \citet{gong-etal-2024-mapping}.

\subsection{Emotion Annotation Finalization}
Following the preliminary emotion annotation process, two annotators who demonstrated the highest quality in utterance transcriptions were selected to provide second-opinion emotion labels for utterances they had not annotated during the initial labelling phase. After this second round of labelling, the final emotion label for each utterance in the dataset was determined through a majority voting procedure. In cases of inter-annotator disagreement regarding the appropriate emotion label, the final decision was made by an external Akan-speaking consultant, recognized as an expert in Akan Emotion Analysis. Notably, an analysis of inter-annotator agreement yielded an overall Fleiss’ Kappa statistic \cite{fleiss1971} of $k = \pmb{0.488}$ which is comparable to the inter-annotator agreement scores of several other popular high-quality speech emotion datasets such as MELD \cite{poria2018meld} which has a score of 0.43 \cite{poria2018meld}, IEMOCAP which has a score of 0.48 \cite{busso2008iemocap}, MSP-IMPROV which has a score of 0.49 \cite{busso2016msp} and M³ED which has a score of 0.59 \cite{zhao-etal-2022-m3ed}.

\begin{table}[t]
  \centering
  \resizebox{0.48\textwidth}{!}{%
  \begin{tabular}{lc}
    \hline
    \textbf{General Statistics} & \textbf{Values}\\
    \hline
    Total number of seconds & 87441\\
    Avg. number of seconds per movie & 4163.4\\
    Total number of movies & 21    \\
    Total number of dialogs     & 385         \\
    Total number of words     &    117305      \\
    Total number of utterances     & 6162          \\
    Total number of turns      & 4477            \\
    Number of prominence words     & 37314           \\
    Number of non-prominence words     & 79991           \\
    Average number of turns per dialog     & 11.62           \\
    Average number of utterances per dialog    & 16           \\
    Average number of words per dialog    &     305       \\
    Average utterance length in seconds     & 6.701           \\
    Average number of words per utterance     &    19        \\
    Average duration per dialog in seconds    & 227.1          \\\hline
  \end{tabular}
  }
  \caption{General statistics of the AkaCE Dataset}
  \label{tab:accents}
\end{table}

\subsection{Prosodic Prominence Annotation}
The annotation strategy used for prosodic prominence closely mirrored the approach employed for emotion labelling. The same two annotators responsible for assigning emotion labels to the utterances were selected for this task. Before starting the prosodic prominence annotations, they received detailed instructions outlining the concept of prosodic prominence and the steps involved in performing the task. Additionally, they were presented with examples of prosodic prominence annotations deemed accurate by consulted linguists to ensure a clear understanding of the expectations.

For the annotation task, the annotators were instructed to listen to the audio of each utterance in the dataset and assign a value of 1 to words they deemed prosodically prominent and 0 to words they considered non-prominent. All annotations were conducted using Excel sheets. This approach to prosodic prominence annotation was inspired by a similar approach leveraged in \citet{cole2017crowd}.

An analysis of the inter-annotator agreement for prosodic annotation between the two annotators yielded an overall Fleiss’ Kappa statistic \cite{fleiss1971} of $k = \pmb{1.0}$. The observed perfect agreement may be attributed to the tonal and expressive nature of Akan, which provides listeners with clear prosodic cues for identifying prominent words. In particular, prior work by \citet{kugler2012prosodic} shows that Akan speakers signal pragmatic prominence through consistent acoustic patterns such as pitch lowering, even on high-toned syllables. This systematic use of prosody to express information structure likely contributes to the high salience of prominent words in speech, facilitating consistent annotation across raters.

\subsection{Dataset Statistics}
\paragraph{General Dataset Statistics}
Table \ref{tab:accents} presents basic statistics of the Akan Cinematic Emotions (AkaCE) Dataset. It contains 385 dialogues, 4477 turns and 6162 utterances, which contain an average of 19 words. With respect to prosodic prominence, 37314 words were annotated to be prosodically prominent whereas 79991 words were annotated to be non-prominent. 

\paragraph{Emotion Distribution}
Table \ref{tab:emotions_dist} illustrates the distribution of emotions in the AkaCE Dataset. Neutral emotion had the highest frequency, appearing in 2,941 instances, while Fear had the lowest frequency, occurring only 134 times.
\begin{table}[t]
  \centering
  \resizebox{0.27\textwidth}{!}{%
  \begin{tabular}{lc}
    \hline
    \textbf{Emotion Labels} & \textbf{Values} \\
    \hline
    Neutral    & 2941           \\
    Sadness     & 806           \\
    Anger   &  1107          \\
    Fear   & 134           \\
    Surprise   & 364           \\
    Disgust   & 162          \\
    Happy   & 568           \\\hline
  \end{tabular}
  }
  \caption{Distribution of emotions in the AkaCE dataset}
  \label{tab:emotions_dist}
\end{table}

\paragraph{Speaker Gender Distribution}
The number of speakers in the AkaCE dataset is 308, of which 155 are men and 153 are women,  as shown in Table \ref{tab:speakers_dist}. 
\begin{table}[t]
\centering
\resizebox{0.32\textwidth}{!}{%
  \begin{tabular}{lc}
    \hline
    \textbf{Speaker statistics} & \textbf{Values} \\
    \hline
    Number of speakers   & 308         \\
    Number of male speakers     & 155  \\
    Number of female speakers     & 153  \\\hline
  \end{tabular}
  }
  \caption{Distribution of speakers in the AkaCE Dataset}
  \label{tab:speakers_dist}
\end{table}

\section{Experiments and Analysis}
We conduct a series of experiments to establish baseline performance for emotion recognition on AkaCE using unimodal and multimodal approaches. We first evaluate text, audio, and vision separately with state-of-the-art models, then explore modality combinations through feature concatenation and transformer-based fusion. These results serve as a foundation for future research on multimodal emotion recognition in Akan.

\subsection{Experiment Setup}
Each movie in our dataset is segmented into training, testing, and validation sets using a 7:1.5:1.5 ratio. Following a comprehensive data cleaning process that removed invalid utterances -- specifically those with erroneous timestamps or annotations -- the final dataset comprised 3,888 utterances for training, 816 for validation, and 834 for testing.

Segmentation for both audio and video modalities is based on the timestamps associated with each utterance. The audio recordings, originally sampled at 44 kHz, are resampled to 16 kHz to meet the input requirements of the Whisper \cite{radford2022robustspeechrecognitionlargescale} model. Video frames are extracted at two distinct rates -- 1 frame per second and 5 frames per second -- to evaluate the impact of temporal resolution on emotion detection. Additionally, MTCNN \cite{Zhang_2016} is employed to extract faces from each frame, capturing crucial facial cues essential for effective emotion recognition.

We conduct our experiments on an RTX A6000 GPU. To ensure a reliable assessment of model performance, we use weighted F1 and macro F1 scores instead of accuracy, as the latter can be misleading in imbalanced scenarios.

\begin{table}[b]
\centering
\resizebox{0.37\textwidth}{!}{%
    \begin{tabular}{lcc}
    \hline
    \textbf{Setting} & \textbf{Weighted F1} & \textbf{Macro F1} \\
    \hline
    No Context       & 43.12               & 18.85            \\
    Context          & \textbf{44.58}               & \textbf{22.29}            \\
    \hline
    \end{tabular}
}
\caption{Text-based emotion detection results using the Ghana-NLP/abena-base-asante-twi-uncased model.}
\label{tab:text_results}
\end{table}

\subsection{Text Experiments}

For our text experiments, we employ the Ghana-NLP/abena-base-asante-twi-uncased \cite{alabi-etal-2020-massive} model from Hugging Face, a variant of multilingual BERT (mBERT) fine-tuned specifically for the Akan language. The model is initially trained on the Twi subset of the JW300 \cite{agic-vulic-2019-jw300} dataset, which primarily consists of the Akuapem dialect of Twi, and is later fine-tuned on Asante Twi Bible data to specialize in Asante Twi. To our knowledge, this remains the only available language model trained on this language.

We investigate the impact of context by comparing two settings: one incorporating the previous utterance as context and another without contextual information. Following the context modeling approach from MMML \cite{wu-etal-2024-multimodal}, we process the context and current utterance separately before concatenating their feature representations, rather than simply merging them at the input level. The concatenated features are then passed to the classifier layer. We use a learning rate of 1e-5 and a batch size of 16 in both settings.

As shown in Table \ref{tab:text_results}, our results indicate that incorporating context improves performance. Specifically, the model without context achieves a weighted F1 score of 43.12 and a macro F1 score of 18.85, while the context-aware model yields a weighted F1 score of 44.58 and a macro F1 score of 22.29. These findings highlight the benefits of incorporating contextual information for emotion detection in Akan text.

\subsection{Audio Experiments}
We conduct audio experiments using three different encoding methods: Whisper\footnote{https://huggingface.co/openai/whisper-small}, spectrogram-based features, and openSMILE\footnote{https://audeering.github.io/opensmile-python/}, where Whisper achieves the highest performance (Table \ref{tab:audio_results}). We set the learning rate to 1e-5 and use a batch size of 16 for all three methods.

OpenSMILE features are extracted using the ComParE 2016 feature set, incorporating Low-Level Descriptors (LLDs) and Functionals, resulting in a 130-dimensional feature vector with a maximum sequence length of 3000. These features are then used to train an audio transformer encoder, following the approach of \citet{wu-etal-2024-multimodal}. 
The model consists of three transformer encoder layers, each with two attention heads, and positional encoding, followed by a fully connected linear classifier for prediction.
The model reaches a weighted F1 score of 13.80 and a macro F1 score 6.58. This low performance can be attributed to the absence of pretraining.

Spectrogram features are computed with 128 Mel-frequency bins, normalized, and truncated or padded to a maximum length of 1024 frames. These features are then used to fine-tune a pretrained Audio Spectrogram Transformer (AST) \cite{gong21b_interspeech} with an additional linear classifier layer. The model achieves a  weighted F1 score of 47.89 and a macro F1 score of 23.36. 
We select AST due to its pretraining on a diverse auditory data, encompassing both human speech and non-human sounds, such as music and environmental noises. This broad training enables AST to effectively capture complex acoustic patterns, making it particularly well-suited for our AkaCE dataset, which consists of movie scenes containing a mix of dialogue, background music, and ambient sounds.

Finally, we fine-tune a Whisper-Small encoder without freezing its parameters, achieving the best performance with a weighted F1 score of 52.38 and a macro F1 score of 29.51. These results indicate that pretraining audio models on multiple languages benefits speech emotion recognition in low-resource target languages. However, due to the imbalance in training samples for Whisper, the improvement remains relatively small. This further underscores the necessity of our dataset collection, as it represents the first multimodal emotion dialogue dataset for an African language, addressing the significant resource gap in emotion recognition research for low-resource languages.

\begin{table}[t]
\centering
\resizebox{0.4\textwidth}{!}{%
    \begin{tabular}{lcc}
    \hline
    \textbf{Model} & \textbf{Weighted F1} & \textbf{Macro F1} \\
    \hline
    openSMILE          & 13.80               & 6.58 \\
    Spectrogram     & 47.89             & 23.36 \\
    Whisper-small       & \textbf{52.38}    & \textbf{29.51} \\
    \hline
    \end{tabular}
}
\caption{Audio-based emotion detection results.}
\label{tab:audio_results}
\end{table}

\subsection{Vision Experiments}

For the vision modality, we explore two main approaches for encoding visual information. First, we use ResNet18 and ResNet50 \cite{he2015deepresiduallearningimage} to extract feature representations from entire video frames. To evaluate the impact of temporal resolution on emotion detection, we experiment with frame sampling rates of 1 frame per second (1 fps) and 5 frames per second (5 fps). In addition, we investigate a face-based approach where faces are extracted from each frame using MTCNN and then encoded with InceptionResnetV1, a model pre-trained on VGGFace2 \cite{cao2018vggface2datasetrecognisingfaces}. All vision experiments are conducted using a learning rate of 1e-4 and a batch size of 16.

\begin{table}[t]
\centering
\resizebox{0.45\textwidth}{!}{%
    \begin{tabular}{lcc}
    \hline
    \textbf{Model} & \textbf{Weighted F1} & \textbf{Macro F1} \\
    \hline
    ResNet18-1fps            & 40.57	     & \textbf{20.02} \\
    ResNet50-1fps            & 38.19	          & 15.1 \\
    ResNet18-5fps            & \textbf{42.04}	     & 17.92 \\
    ResNet50-5fps            & 41.76	       & 19 \\
    Inception-Face-5fps  & 39.96	   & 16.53\\
    \hline
    \end{tabular}
}
\caption{Vision-based emotion detection results.}
\label{tab:vision_results}
\end{table}

As shown in Table \ref{tab:vision_results}, our results indicate that ResNet18 with a 5 fps sampling rate achieves the highest weighted F1 score (42.04), suggesting that increasing temporal resolution enhances emotion recognition. However, the highest macro F1 score (20.02) is observed with ResNet18 at 1 fps, indicating that this setting may better capture underrepresented emotion classes. Interestingly, ResNet50, despite being a larger model, does not consistently outperform ResNet18, possibly due to overfitting. Its best weighted F1 score (41.76 at 5 fps) slightly trails that of ResNet18-5fps.

The face-based approach using InceptionResNetV1 underperforms compared to whole-frame models, achieving only 39.96 weighted F1 and 16.53 macro F1, suggesting that facial expressions alone may not provide sufficient information for robust emotion detection in our dataset. Unlike datasets such as CMU-MOSEI \cite{zadeh2018multimodal} that enforce a single visible face in close-up shots, our dataset does not impose such constraints. As a result, videos may contain multiple faces, and the primary speaker’s face may be distant from the camera, adding challenges for models relying solely on facial features.  These findings highlight the importance of frame selection strategies and suggest that balancing temporal resolution with model capacity is crucial for optimal vision-based emotion recognition.

\begin{figure*}[ht]
    \centering
    \includegraphics[width=0.8\textwidth]{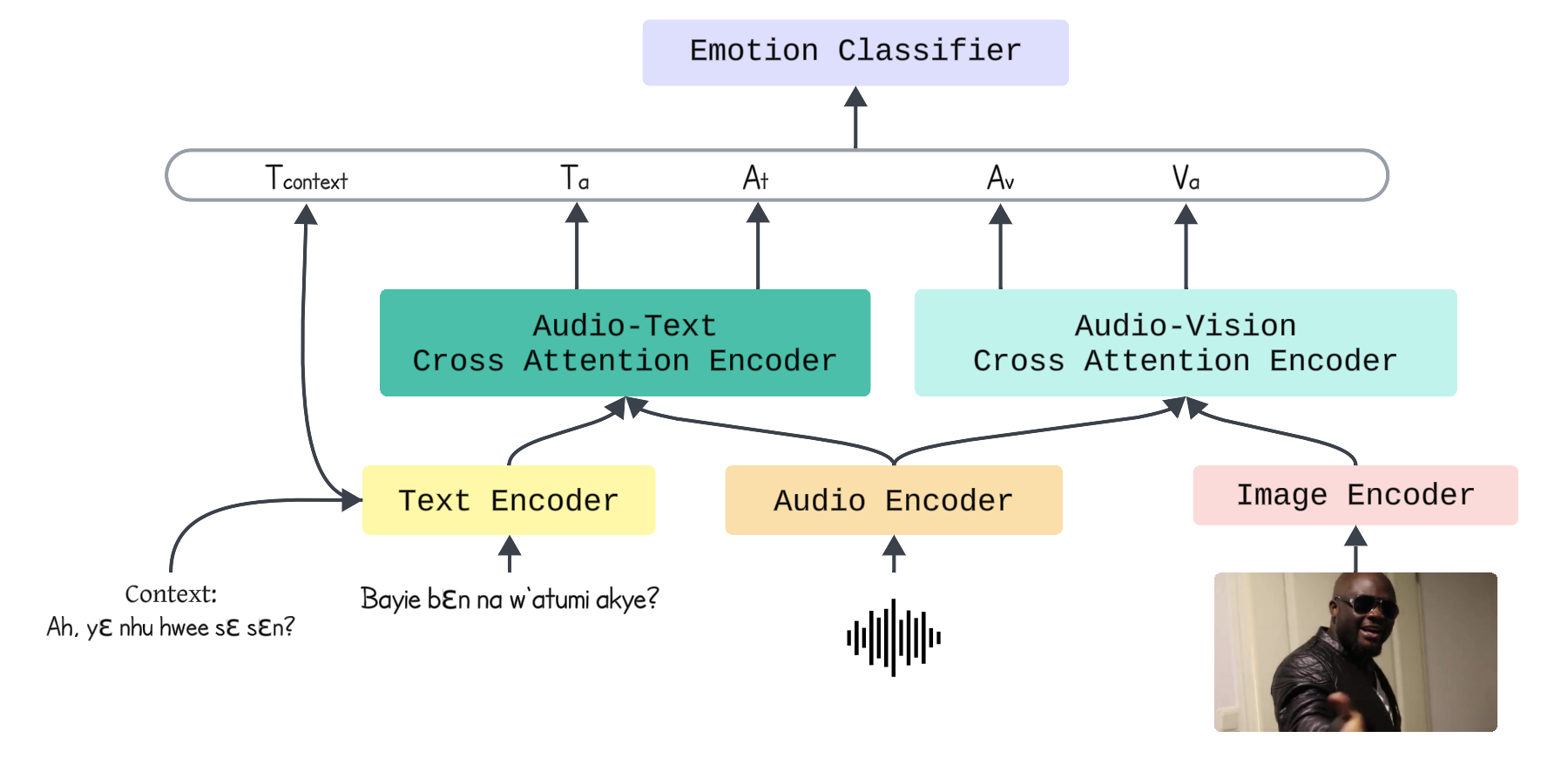}
    \caption{Illustration of the transformer fusion model.}
    \label{fig:label}
\end{figure*}

\subsection{Multimodal Experiments}
We evaluate modality combinations using the best-performing unimodal models. Starting with simple feature concatenation as a baseline, we then apply transformer-based fusion to enhance cross-modal interactions. These experiments assess the impact of multimodal integration on emotion recognition.

\label{sec:fusion}
\subsubsection{Modality Features Concatenation}
For the fusion experiments, we evaluate all possible combinations of the three modalities to understand how multimodal integration impacts ERC. We use the best-performing unimodal models: the contextual text model (Ghana-NLP/abena-base-asante-twi-uncased) for text, Whisper-small for audio, and ResNet18-5fps for vision. Feature representations from each modality are concatenated and passed through a classifier layer to compute logits for emotion prediction. To consider distinct characteristics of each modality, we experiment with different learning rates but find that using a single learning rate of 1e-5 yields the most stable results.

\begin{table}[t]
\centering
\resizebox{0.47\textwidth}{!}{%
    \begin{tabular}{lcc}
    \hline
    \textbf{Modality} & \textbf{Weighted F1} & \textbf{Macro F1} \\
    \hline
    Text                        & 44.58 & 22.29 \\
    Audio                       & 52.38	 & 29.51 \\
    Vision                      & 40.57 & 20.02 \\
    Text + Audio                & 55.51	& 30.15 \\
    Text + Vision               & 43.33	 & 21.15 \\
    Audio + Vision              & 53.84 & 30.42 \\
    Text + Audio + Vision       & \textbf{55.81}	 & \textbf{30.97} \\
    \hline
    \end{tabular}
}
\caption{Results of modality concatenation experiments using the best unimodal models.}
\label{tab:modality_results}
\end{table}

Our results in Table \ref{tab:modality_results} show that combining modalities improves emotion recognition performance, with the best results achieved when integrating all three modalities. The multimodal model using text, audio, and vision achieves the highest weighted F1 (55.81) and macro F1 (30.97), outperforming all unimodal and bimodal models. 

Among the unimodal models, audio performs best (52.38 weighted F1, 29.51 macro F1), indicating that speech features carry the most discriminative information for emotion recognition in our dataset. Interestingly, text alone (44.58 weighted F1, 22.29 macro F1) underperforms compared to audio, contrasting with trends in high-resource languages where text embeddings often yield the best results \cite{zadeh2018multimodal,yu2020ch}. This gap is likely due to the limited availability of large and diverse pretraining corpora for Akan, restricting the effectiveness of text embeddings. Vision alone performs worst (40.57 weighted F1), suggesting that visual cues are less reliable, possibly due to variations in facial visibility and camera angles.

In bimodal settings, text + audio (55.51 weighted F1) and audio + vision (53.84 weighted F1) show substantial improvements over their unimodal counterparts, reinforcing the importance of speech information in multimodal emotion recognition. However, text + vision (43.33 weighted F1) provides only a marginal improvement over vision alone, suggesting that textual and visual features may not be as complementary as text and audio. Overall, these results highlight the advantages of multimodal fusion, particularly the strong synergy between textual and auditory features, while also emphasizing the challenges posed by the limited availability of high-quality pretraining data for Akan.

\subsubsection{Transformer Fusion}
To further enhance multimodal fusion, we employ a transformer-based cross-attention encoder to capture interdependencies between different modalities. This approach enables a more nuanced integration of modality-specific features by projecting information from one modality into the representational space of another. Given our bimodal results indicating that text and vision do not complement each other effectively, we structure our fusion process around audio-centric interactions. Specifically, we use a cross-attention encoder to fuse text and audio (audio-text fusion) as well as audio and vision (audio-vision fusion).

In this framework, we first extract features from each modality using the best-performing unimodal models. The cross-attention encoder is designed such that the query comes from one modality while the keys and values are derived from another. This mechanism allows each modality to selectively attend to the most relevant aspects of the other, facilitating effective multimodal alignment. The encoder projects the hidden representations of one modality into the representational space of another, enhancing cross-modal interactions.

To structure the fusion process, we prepend a CLS token to the hidden states of each modality before applying the cross-attention mechanism. After audio-text fusion, we obtain two new hidden representations: $T_{a}$, where text features are projected into the audio space, and $A_{t}$, where audio features are projected into the text space. Similarly, for audio-vision fusion, we obtain $A_{v}$ and $V_{a}$, corresponding to audio projected into the vision space and vice versa. For classification, we use the CLS token from each fused representation as features. Additionally, we incorporate the context feature from the text encoder to enrich the final representation. These features are concatenated and passed through a classifier layer to predict emotion labels.

\begin{table}[t]
\centering
\resizebox{0.45\textwidth}{!}{%
    \begin{tabular}{lcc}
    \hline
    \textbf{Model} & \textbf{Weighted F1} & \textbf{Macro F1} \\
    \hline
    Concatenation           & 55.81     & 30.97\\
    Transformer Fusion            & \textbf{56.13}         & \textbf{31.68} \\
    
    \hline
    \end{tabular}
}
\caption{Results of multimodal fusion experiments.}
\label{tab:fusion_results}
\end{table}

\begin{table}[t]
\centering
\resizebox{0.35\textwidth}{!}{%
    \begin{tabular}{lcc}
    \hline
    \textbf{Emotion Class} & \textbf{F1} & \textbf{Support} \\
    \hline
    Neutral         & 68.57     & 409\\
    Anger           & 51.50     & 176 \\
    Sadness         & 56.48     & 128 \\
    Joy             & 25.32     & 68 \\
    Surprise        & 15.21     & 26 \\
    Fear            & 0.08      & 16 \\
    Disgust         & 0.02      & 11 \\
    
    \hline
    \end{tabular}
}
\caption{Emotion-wide results of multimodal fusion experiments.}
\label{tab:emotion_wide_fusion_results}
\end{table}

As shown in Table \ref{tab:fusion_results}, Transformer fusion outperforms simple concatenation in both weighted and macro F1 scores, achieving 56.13 and 31.68, respectively. This improvement highlights the effectiveness of advanced fusion mechanisms in integrating multimodal features for emotion recognition. The higher macro F1 score suggests that Transformer Fusion provides better balance across emotion classes, likely due to its ability to capture cross-modal dependencies more effectively. These findings underscore the potential of attention-based fusion techniques for enhancing multimodal ERC, particularly in low-resource settings like Akan. 

Table~\ref{tab:emotion_wide_fusion_results} presents the per-emotion F1 scores and support. As expected, the model performs best on classes with higher frequency in the training set, such as \textit{Neutral}, \textit{Anger}, and \textit{Sadness}, while performance on low-resource classes like \textit{Disgust} and \textit{Fear} remains limited. These findings underscore the importance of addressing class imbalance in multimodal emotion recognition, especially in low-resource language contexts such as Akan.

\paragraph{Model Complexity.}
To contextualize the scale of our proposed transformer fusion system, we report in Table~\ref{tab:model_params} the parameter counts for each of its core components. For the vision backbone, we use ResNet-18 (convolutional and fully connected layers), totalling approximately 11.7 million parameters. Text features are extracted using the Ghana-NLP \texttt{abena-base-asante-twi-uncased} model, a BERT-base variant with 12 transformer layers of 768 hidden units, contributing about 178 million parameters. Audio features are encoded with OpenAI’s Whisper-small model, which consists of 12-layer encoder-decoder stacks with 768-dimensional hidden representations, accounting for 244 million parameters. The transformer fusion module itself is composed of five cross-attention blocks, each using 768-dimensional queries, keys, and values with 12 attention heads, adding roughly 12 million parameters. Altogether, the complete model comprises approximately \textbf{450 million parameters}, with the majority allocated to the unimodal backbones and only a small fraction to the fusion mechanism.

\begin{table}[h]
\centering
\resizebox{0.45\textwidth}{!}{%
    \begin{tabular}{l c c}
    \hline
    \textbf{Component} & \textbf{Configuration} & \textbf{Parameters} \\
    \hline
    ResNet-18 (Vision) & Conv + FC layers & 11.7M \\
    BERT-base (Text) & 12×768 (Abena-Twi) & 178M \\
    Whisper-small (Audio) & 12×768 Encoder-Decoder & 244M \\
    Cross-Attention Blocks & 5×(768-d, 12 heads) & 12M \\
    \hline
    \textbf{Total} & & \textbf{$\approx 450M$} \\
    \hline
    \end{tabular}
}
\caption{Model parameter breakdown across modalities.}
\label{tab:model_params}
\end{table}

\section{Conclusion and Future Directions}
We introduce the Akan Conversation Emotion (AkaCE) dataset, the first multimodal emotion dialogue dataset for an African language, addressing the resource gap in ERC research for low-resource languages. AkaCE comprises 385 emotion-labeled dialogues and word-level prosodic prominence annotations, making it a valuable resource for cross-cultural emotion recognition and tonal language prosody research. Our experiments with state-of-the-art ERC methods validate AkaCE’s quality and establish a strong baseline for future research.


Looking ahead, we aim to expand to additional African languages and develop culturally adaptive ERC systems. Multimodal emotion recognition can be improved by speech enhancement techniques and pretraining models on African languages. Integrating vision-language models for scene descriptions can also provide richer context. Advanced fusion techniques like graph neural networks (GNNs) and hypergraphs may further refine cross-modal integration. We hope AkaCE inspires further research toward culturally adaptive, linguistically diverse NLP resources.
\section*{Acknowledgements}
This research is supported in part by the National Science Foundation via ARNI (The NSF AI Institute for Artificial and Natural Intelligence), under the Columbia 2025 Research Project ("Towards Safe, Robust, Interpretable Dialogue Agents for Democratized Medical Care"), and in part by the Defense
Advanced Research Projects Agency (DARPA), via
the CCU Program contract HR001122C0034. The views, opinions and/or findings expressed are those
of the authors and should not be interpreted as
representing the official views or policies of the National
Science Foundation or the U.S. Government.

\section*{Limitations}

While the Akan Cinematic Emotions (AkaCE) dataset represents a significant advancement in multimodal emotion recognition research, particularly for African languages, there are several limitations to acknowledge.

One limitation of this work is that the dataset focuses exclusively on the Akan language. While this contributes to the representation of low-resource languages in emotion recognition research, the findings may not generalize to other African languages or cultural contexts without further adaptation and testing. The emotional expressions and prosodic characteristics in Akan may differ substantially from those in other languages, limiting cross-linguistic applicability.

Another limitation lies in the domain of the dataset, which is derived from movie dialogues. While this ensures the presence of diverse emotions and rich multimodal interactions, it is likely that a portion of the data contains acted emotions rather than naturally occurring ones. Acted emotions may differ in intensity, expression, and prosodic features from emotions encountered in real-world scenarios, potentially introducing a bias in models trained on this dataset. This could affect the generalizability of such models to real-life applications, where emotional expressions might be less exaggerated or contextually different.

Additionally, while the inclusion of prosodic annotations is a novel feature, the labelling process may be subject to subjective interpretations, particularly for ambiguous emotional expressions. The quality and consistency of these annotations could impact the performance of models relying on prosodic features. Further efforts to standardize prosodic annotation practices would benefit future iterations of this dataset.

Another challenge is related to visual data. Although the dataset incorporates visual modalities, the quality and consistency of visual features in movie dialogues may vary due to differences in lighting, camera angles, and actor positioning. These variations could impact the reliability of visual emotion recognition models trained on this dataset. Moreover, further exploration of vision features, including fine-tuned embeddings and advanced visual annotations, may reveal additional insights but was not the focus of this study.

Despite these limitations, we believe that AkaCE provides an essential foundation for advancing speech emotion recognition in low-resource languages and encourages further exploration in this area.

\section*{Ethical Considerations}
The potential for misuse of the AkaCE dataset must be carefully acknowledged. While the dataset is intended for research purposes, deploying models trained on AkaCE in real-world applications without proper domain adaptation and validation could result in inaccurate emotion predictions, particularly in scenarios that deviate from cinematic dialogues. As such, researchers and practitioners should exercise caution when extending the use of this dataset to other applications.

\bibliography{anthology,custom}

\begin{thebibliography}{53}
\providecommand{\natexlab}[1]{#1}

\bibitem[{Agi{\'c} and Vuli{\'c}(2019)}]{agic-vulic-2019-jw300}
{\v{Z}}eljko Agi{\'c} and Ivan Vuli{\'c}. 2019.
\newblock \href {https://doi.org/10.18653/v1/P19-1310} {{JW}300: A wide-coverage parallel corpus for low-resource languages}.
\newblock In \emph{Proceedings of the 57th Annual Meeting of the Association for Computational Linguistics}, pages 3204--3210, Florence, Italy. Association for Computational Linguistics.

\bibitem[{Ahmad et~al.(2025)Ahmad, Dudy, Belay, Abdulmumin, Yimam, Muhammad, and Church}]{ahmad2025exploring}
Ibrahim~Said Ahmad, Shiran Dudy, Tadesse~Destaw Belay, Idris Abdulmumin, Seid~Muhie Yimam, Shamsuddeen~Hassan Muhammad, and Kenneth Church. 2025.
\newblock Exploring cultural nuances in emotion perception across 15 african languages.
\newblock \emph{arXiv preprint arXiv:2503.19642}.

\bibitem[{Alabi et~al.(2020)Alabi, Amponsah-Kaakyire, Adelani, and Espa{\~n}a-Bonet}]{alabi-etal-2020-massive}
Jesujoba~O. Alabi, Kwabena Amponsah-Kaakyire, David~I. Adelani, and Cristina Espa{\~n}a-Bonet. 2020.
\newblock \href {https://aclanthology.org/2020.lrec-1.335/} {Massive vs. curated embeddings for low-resourced languages: the case of {Y}or{\`u}b{\'a} and {T}wi}.
\newblock In \emph{Proceedings of the Twelfth Language Resources and Evaluation Conference}, pages 2754--2762, Marseille, France. European Language Resources Association.

\bibitem[{Busso et~al.(2008)Busso, Bulut, Lee, Kazemzadeh, Mower, Kim, Chang, Lee, and Narayanan}]{busso2008iemocap}
Carlos Busso, Murtaza Bulut, Chi-Chun Lee, Abe Kazemzadeh, Emily Mower, Samuel Kim, Jeannette~N Chang, Sungbok Lee, and Shrikanth~S Narayanan. 2008.
\newblock Iemocap: Interactive emotional dyadic motion capture database.
\newblock \emph{Language resources and evaluation}, 42:335--359.

\bibitem[{Busso et~al.(2016)Busso, Parthasarathy, Burmania, AbdelWahab, Sadoughi, and Provost}]{busso2016msp}
Carlos Busso, Srinivas Parthasarathy, Alec Burmania, Mohammed AbdelWahab, Najmeh Sadoughi, and Emily~Mower Provost. 2016.
\newblock Msp-improv: An acted corpus of dyadic interactions to study emotion perception.
\newblock \emph{IEEE Transactions on Affective Computing}, 8(1):67--80.

\bibitem[{Cao et~al.(2018)Cao, Shen, Xie, Parkhi, and Zisserman}]{cao2018vggface2datasetrecognisingfaces}
Qiong Cao, Li~Shen, Weidi Xie, Omkar~M. Parkhi, and Andrew Zisserman. 2018.
\newblock \href {https://arxiv.org/abs/1710.08092} {Vggface2: A dataset for recognising faces across pose and age}.
\newblock \emph{Preprint}, arXiv:1710.08092.

\bibitem[{Chen et~al.(2018)Chen, Hsu, Kuo, and Ku}]{chen1802emotionlines}
SY~Chen, CC~Hsu, CC~Kuo, and LW~Ku. 2018.
\newblock Emotionlines: An emotion corpus of multi-party conversations. arxiv 2018.
\newblock \emph{arXiv preprint arXiv:1802.08379}.

\bibitem[{Chudasama et~al.(2022)Chudasama, Kar, Gudmalwar, Shah, Wasnik, and Onoe}]{chudasama2022m2fnet}
Vishal Chudasama, Purbayan Kar, Ashish Gudmalwar, Nirmesh Shah, Pankaj Wasnik, and Naoyuki Onoe. 2022.
\newblock M2fnet: Multi-modal fusion network for emotion recognition in conversation.
\newblock In \emph{Proceedings of the IEEE/CVF Conference on Computer Vision and Pattern Recognition}, pages 4652--4661.

\bibitem[{Cole et~al.(2017)Cole, Mahrt, and Roy}]{cole2017crowd}
Jennifer Cole, Timothy Mahrt, and Joseph Roy. 2017.
\newblock Crowd-sourcing prosodic annotation.
\newblock \emph{Computer Speech \& Language}, 45:300--325.

\bibitem[{Danieli et~al.(2015)Danieli, Riccardi, and Alam}]{danieli2015emotion}
Morena Danieli, Giuseppe Riccardi, and Firoj Alam. 2015.
\newblock Emotion unfolding and affective scenes: A case study in spoken conversations.
\newblock In \emph{Proceedings of the International Workshop on Emotion Representations and Modelling for Companion Technologies}, pages 5--11.

\bibitem[{Dhall et~al.(2012)Dhall, Goecke, Lucey, and Gedoen}]{fromcollecting}
Abhinav Dhall, Roland Goecke, Simon Lucey, and Tom Gedoen. 2012.
\newblock Collecting large, richly annotated facial-expression databases from movies.
\newblock \emph{IEEE Multimedia}, pages 34--41.

\bibitem[{Dossou(2024)}]{dossou2024advancingafricanaccentedspeechrecognition}
Bonaventure F.~P. Dossou. 2024.
\newblock \href {https://arxiv.org/abs/2306.02105} {Advancing african-accented speech recognition: Epistemic uncertainty-driven data selection for generalizable asr models}.
\newblock \emph{Preprint}, arXiv:2306.02105.

\bibitem[{Ekman(1992)}]{ekman1992there}
Paul Ekman. 1992.
\newblock Are there basic emotions?
\newblock \emph{Psychological review}, 99 (3).

\bibitem[{Fleiss(1971)}]{fleiss1971}
Joseph~L. Fleiss. 1971.
\newblock \href {https://doi.org/10.1037/h0031619} {Measuring nominal scale agreement among many raters}.
\newblock \emph{Psychological Bulletin}, 76(5):378--382.

\bibitem[{Fragopanagos and Taylor(2005)}]{fragopanagos2005emotion}
Nickolaos Fragopanagos and John~G Taylor. 2005.
\newblock Emotion recognition in human--computer interaction.
\newblock \emph{Neural Networks}, 18(4):389--405.

\bibitem[{Fu(2024)}]{fu2024ckerc}
Yumeng Fu. 2024.
\newblock Ckerc: Joint large language models with commonsense knowledge for emotion recognition in conversation.
\newblock \emph{arXiv preprint arXiv:2403.07260}.

\bibitem[{Ghosal et~al.(2019)Ghosal, Majumder, Poria, Chhaya, and Gelbukh}]{ghosal2019dialoguegcn}
Deepanway Ghosal, Navonil Majumder, Soujanya Poria, Niyati Chhaya, and Alexander Gelbukh. 2019.
\newblock Dialoguegcn: A graph convolutional neural network for emotion recognition in conversation.
\newblock \emph{arXiv preprint arXiv:1908.11540}.

\bibitem[{Gong et~al.(2021)Gong, Chung, and Glass}]{gong21b_interspeech}
Yuan Gong, Yu-An Chung, and James Glass. 2021.
\newblock \href {https://doi.org/10.21437/Interspeech.2021-698} {Ast: Audio spectrogram transformer}.
\newblock In \emph{Interspeech 2021}, pages 571--575.

\bibitem[{Gong et~al.(2024)Gong, Yao, Hu, Zhu, and Hirschberg}]{gong-etal-2024-mapping}
Ziwei Gong, Muyin Yao, Xinyi Hu, Xiaoning Zhu, and Julia Hirschberg. 2024.
\newblock \href {https://aclanthology.org/2024.law-1.3/} {A mapping on current classifying categories of emotions used in multimodal models for emotion recognition}.
\newblock In \emph{Proceedings of The 18th Linguistic Annotation Workshop (LAW-XVIII)}, pages 19--28, St. Julians, Malta. Association for Computational Linguistics.

\bibitem[{Hazarika et~al.(2018{\natexlab{a}})Hazarika, Poria, Mihalcea, Cambria, and Zimmermann}]{hazarika-etal-2018-icon}
Devamanyu Hazarika, Soujanya Poria, Rada Mihalcea, Erik Cambria, and Roger Zimmermann. 2018{\natexlab{a}}.
\newblock \href {https://doi.org/10.18653/v1/D18-1280} {{ICON}: Interactive conversational memory network for multimodal emotion detection}.
\newblock In \emph{Proceedings of the 2018 Conference on Empirical Methods in Natural Language Processing}, pages 2594--2604, Brussels, Belgium. Association for Computational Linguistics.

\bibitem[{Hazarika et~al.(2018{\natexlab{b}})Hazarika, Poria, Zadeh, Cambria, Morency, and Zimmermann}]{hazarika2018conversational}
Devamanyu Hazarika, Soujanya Poria, Amir Zadeh, Erik Cambria, Louis-Philippe Morency, and Roger Zimmermann. 2018{\natexlab{b}}.
\newblock Conversational memory network for emotion recognition in dyadic dialogue videos.
\newblock In \emph{Proceedings of the conference. Association for Computational Linguistics. North American Chapter. Meeting}, volume 2018, page 2122. NIH Public Access.

\bibitem[{He et~al.(2015)He, Zhang, Ren, and Sun}]{he2015deepresiduallearningimage}
Kaiming He, Xiangyu Zhang, Shaoqing Ren, and Jian Sun. 2015.
\newblock \href {https://arxiv.org/abs/1512.03385} {Deep residual learning for image recognition}.
\newblock \emph{Preprint}, arXiv:1512.03385.

\bibitem[{Hu et~al.(2021)Hu, Liu, Zhao, and Jin}]{hu2021mmgcn}
Jingwen Hu, Yuchen Liu, Jinming Zhao, and Qin Jin. 2021.
\newblock Mmgcn: Multimodal fusion via deep graph convolution network for emotion recognition in conversation.
\newblock \emph{arXiv preprint arXiv:2107.06779}.

\bibitem[{Jiao et~al.(2019)Jiao, Yang, King, and Lyu}]{jiao2019higru}
Wenxiang Jiao, Haiqin Yang, Irwin King, and Michael~R Lyu. 2019.
\newblock Higru: Hierarchical gated recurrent units for utterance-level emotion recognition.
\newblock \emph{arXiv preprint arXiv:1904.04446}.

\bibitem[{Kim and Vossen(2021)}]{kim2021emoberta}
Taewoon Kim and Piek Vossen. 2021.
\newblock Emoberta: Speaker-aware emotion recognition in conversation with roberta.
\newblock \emph{arXiv preprint arXiv:2108.12009}.

\bibitem[{K{\"u}gler and Genzel(2012)}]{kugler2012prosodic}
Frank K{\"u}gler and Susanne Genzel. 2012.
\newblock On the prosodic expression of pragmatic prominence: The case of pitch register lowering in akan.
\newblock \emph{Language and speech}, 55(3):331--359.

\bibitem[{Leben(2018)}]{leben2018languages}
William~R Leben. 2018.
\newblock Languages of the world.
\newblock In \emph{Oxford Research Encyclopedia of Linguistics}.

\bibitem[{Lei et~al.(2023)Lei, Dong, Wang, Wang, and Wang}]{lei2023instructerc}
Shanglin Lei, Guanting Dong, Xiaoping Wang, Keheng Wang, and Sirui Wang. 2023.
\newblock Instructerc: Reforming emotion recognition in conversation with a retrieval multi-task llms framework.
\newblock \emph{arXiv preprint arXiv:2309.11911}.

\bibitem[{Li et~al.(2018)Li, Tao, Schuller, Shan, Jiang, and Jia}]{li2018mec}
Ya~Li, Jianhua Tao, Bj{\"o}rn Schuller, Shiguang Shan, Dongmei Jiang, and Jia Jia. 2018.
\newblock Mec 2017: Multimodal emotion recognition challenge.
\newblock In \emph{2018 First Asian Conference on Affective Computing and Intelligent Interaction (ACII Asia)}, pages 1--5. IEEE.

\bibitem[{Li et~al.(2017)Li, Su, Shen, Li, Cao, and Niu}]{li2017dailydialog}
Yanran Li, Hui Su, Xiaoyu Shen, Wenjie Li, Ziqiang Cao, and Shuzi Niu. 2017.
\newblock Dailydialog: A manually labelled multi-turn dialogue dataset.
\newblock \emph{arXiv preprint arXiv:1710.03957}.

\bibitem[{Majumder et~al.(2019)Majumder, Poria, Hazarika, Mihalcea, Gelbukh, and DialogueRNN}]{majumder2019attentive}
N~Majumder, S~Poria, D~Hazarika, R~Mihalcea, A~Gelbukh, and E~Cambria DialogueRNN. 2019.
\newblock An attentive rnn for emotion detection in conversations.
\newblock \emph{Association for the Advancement of Artificial Intelligence}, pages 6818--6825.

\bibitem[{{Mo Ibrahim Foundation}(2023)}]{moibrahim2023facts}
{Mo Ibrahim Foundation}. 2023.
\newblock \href {https://mo.ibrahim.foundation/sites/default/files/2023-04/2023-facts-figures-global-africa.pdf} {{Africa in the World and the World in Africa: Facts \& Figures, April 2023}}.
\newblock Accessed: 15-Feb-2025.

\bibitem[{Peterson et~al.(n.d.)Peterson, Al-Saleh, Allen, Fochios, Mulford, Paulson-Smith, and Marino}]{lctlresources}
Angeline Peterson, Danya Al-Saleh, Sam Allen, Alex Fochios, Olivia Mulford, Kaden Paulson-Smith, and Lauren~Parnell Marino. n.d.
\newblock \href {https://wisc.pb.unizin.org/lctlresources/front-matter/introduction/} {\emph{Resources for Self-Instructional Learners of Less Commonly Taught Languages}}.
\newblock Accessed: 15-Feb-2025.

\bibitem[{Poria et~al.(2017)Poria, Cambria, Hazarika, Majumder, Zadeh, and Morency}]{poria2017context}
Soujanya Poria, Erik Cambria, Devamanyu Hazarika, Navonil Majumder, Amir Zadeh, and Louis-Philippe Morency. 2017.
\newblock Context-dependent sentiment analysis in user-generated videos.
\newblock In \emph{Proceedings of the 55th annual meeting of the association for computational linguistics (volume 1: Long papers)}, pages 873--883.

\bibitem[{Poria et~al.(2018)Poria, Hazarika, Majumder, Naik, Cambria, and Mihalcea}]{poria2018meld}
Soujanya Poria, Devamanyu Hazarika, Navonil Majumder, Gautam Naik, Erik Cambria, and Rada Mihalcea. 2018.
\newblock Meld: A multimodal multi-party dataset for emotion recognition in conversations.
\newblock \emph{arXiv preprint arXiv:1810.02508}.

\bibitem[{Poria et~al.(2019)Poria, Majumder, Mihalcea, and Hovy}]{poria2019emotion}
Soujanya Poria, Navonil Majumder, Rada Mihalcea, and Eduard Hovy. 2019.
\newblock Emotion recognition in conversation: Research challenges, datasets, and recent advances.
\newblock \emph{IEEE access}, 7:100943--100953.

\bibitem[{Radford et~al.(2022)Radford, Kim, Xu, Brockman, McLeavey, and Sutskever}]{radford2022robustspeechrecognitionlargescale}
Alec Radford, Jong~Wook Kim, Tao Xu, Greg Brockman, Christine McLeavey, and Ilya Sutskever. 2022.
\newblock \href {https://arxiv.org/abs/2212.04356} {Robust speech recognition via large-scale weak supervision}.
\newblock \emph{Preprint}, arXiv:2212.04356.

\bibitem[{Ringeval et~al.(2018)Ringeval, Schuller, Valstar, Cowie, Kaya, Schmitt, Amiriparian, Cummins, Lalanne, Michaud et~al.}]{ringeval2018avec}
Fabien Ringeval, Bj{\"o}rn Schuller, Michel Valstar, Roddy Cowie, Heysem Kaya, Maximilian Schmitt, Shahin Amiriparian, Nicholas Cummins, Denis Lalanne, Adrien Michaud, et~al. 2018.
\newblock Avec 2018 workshop and challenge: Bipolar disorder and cross-cultural affect recognition.
\newblock In \emph{Proceedings of the 2018 on audio/visual emotion challenge and workshop}, pages 3--13.

\bibitem[{Shen et~al.(2020)Shen, Wang, Duan, Li, and Zhu}]{10.1145/3394171.3413909}
Guangyao Shen, Xin Wang, Xuguang Duan, Hongzhi Li, and Wenwu Zhu. 2020.
\newblock \href {https://doi.org/10.1145/3394171.3413909} {Memor: A dataset for multimodal emotion reasoning in videos}.
\newblock In \emph{Proceedings of the 28th ACM International Conference on Multimedia}, MM '20, page 493–502, New York, NY, USA. Association for Computing Machinery.

\bibitem[{Shen et~al.(2021)Shen, Chen, Quan, and Xie}]{shen2021dialogxl}
Weizhou Shen, Junqing Chen, Xiaojun Quan, and Zhixian Xie. 2021.
\newblock Dialogxl: All-in-one xlnet for multi-party conversation emotion recognition.
\newblock In \emph{Proceedings of the AAAI Conference on Artificial Intelligence}, volume~35, pages 13789--13797.

\bibitem[{Tsai et~al.(2019)Tsai, Bai, Liang, Kolter, Morency, and Salakhutdinov}]{tsai2019multimodal}
Yao-Hung~Hubert Tsai, Shaojie Bai, Paul~Pu Liang, J~Zico Kolter, Louis-Philippe Morency, and Ruslan Salakhutdinov. 2019.
\newblock Multimodal transformer for unaligned multimodal language sequences.
\newblock In \emph{Proceedings of the conference. Association for computational linguistics. Meeting}, volume 2019, page 6558. NIH Public Access.

\bibitem[{Wu et~al.(2024{\natexlab{a}})Wu, Gong, Ai, Shi, Donbekci, and Hirschberg}]{wu2024beyond}
Zehui Wu, Ziwei Gong, Lin Ai, Pengyuan Shi, Kaan Donbekci, and Julia Hirschberg. 2024{\natexlab{a}}.
\newblock Beyond silent letters: Amplifying llms in emotion recognition with vocal nuances.
\newblock \emph{arXiv preprint arXiv:2407.21315}.

\bibitem[{Wu et~al.(2024{\natexlab{b}})Wu, Gong, Koo, and Hirschberg}]{wu-etal-2024-multimodal}
Zehui Wu, Ziwei Gong, Jaywon Koo, and Julia Hirschberg. 2024{\natexlab{b}}.
\newblock \href {https://doi.org/10.18653/v1/2024.naacl-long.197} {Multimodal multi-loss fusion network for sentiment analysis}.
\newblock In \emph{Proceedings of the 2024 Conference of the North American Chapter of the Association for Computational Linguistics: Human Language Technologies (Volume 1: Long Papers)}, pages 3588--3602, Mexico City, Mexico. Association for Computational Linguistics.

\bibitem[{Xue et~al.(2024)Xue, Nguyen, Matheny, and Nguyen}]{xue2024bioserc}
Jieying Xue, Minh-Phuong Nguyen, Blake Matheny, and Le-Minh Nguyen. 2024.
\newblock Bioserc: Integrating biography speakers supported by llms for erc tasks.
\newblock In \emph{International Conference on Artificial Neural Networks}, pages 277--292.

\bibitem[{Yu et~al.(2020)Yu, Xu, Meng, Zhu, Ma, Wu, Zou, and Yang}]{yu2020ch}
Wenmeng Yu, Hua Xu, Fanyang Meng, Yilin Zhu, Yixiao Ma, Jiele Wu, Jiyun Zou, and Kaicheng Yang. 2020.
\newblock Ch-sims: A chinese multimodal sentiment analysis dataset with fine-grained annotation of modality.
\newblock In \emph{Proceedings of the 58th annual meeting of the association for computational linguistics}, pages 3718--3727.

\bibitem[{Zadeh et~al.(2018)Zadeh, Liang, Poria, Cambria, and Morency}]{zadeh2018multimodal}
AmirAli~Bagher Zadeh, Paul~Pu Liang, Soujanya Poria, Erik Cambria, and Louis-Philippe Morency. 2018.
\newblock Multimodal language analysis in the wild: Cmu-mosei dataset and interpretable dynamic fusion graph.
\newblock In \emph{Proceedings of the 56th Annual Meeting of the Association for Computational Linguistics (Volume 1: Long Papers)}, pages 2236--2246.

\bibitem[{Zahiri and Choi(2018)}]{zahiri2018emotion}
Sayyed~M Zahiri and Jinho~D Choi. 2018.
\newblock Emotion detection on tv show transcripts with sequence-based convolutional neural networks.
\newblock In \emph{Workshops at the thirty-second aaai conference on artificial intelligence}.

\bibitem[{Zhang et~al.(2019)Zhang, Wu, Sun, Li, Zhu, and Zhou}]{zhang2019modeling}
Dong Zhang, Liangqing Wu, Changlong Sun, Shoushan Li, Qiaoming Zhu, and Guodong Zhou. 2019.
\newblock Modeling both context-and speaker-sensitive dependence for emotion detection in multi-speaker conversations.
\newblock In \emph{IJCAI}, pages 5415--5421. Macao.

\bibitem[{Zhang et~al.(2016)Zhang, Zhang, Li, and Qiao}]{Zhang_2016}
Kaipeng Zhang, Zhanpeng Zhang, Zhifeng Li, and Yu~Qiao. 2016.
\newblock \href {https://doi.org/10.1109/lsp.2016.2603342} {Joint face detection and alignment using multitask cascaded convolutional networks}.
\newblock \emph{IEEE Signal Processing Letters}, 23(10):1499–1503.

\bibitem[{Zhang et~al.(2023)Zhang, Wang, Tiwari, Li, Wang, and Qin}]{zhang2023dialoguellm}
Yazhou Zhang, Mengyao Wang, Prayag Tiwari, Qiuchi Li, Benyou Wang, and Jing Qin. 2023.
\newblock Dialoguellm: Context and emotion knowledge-tuned llama models for emotion recognition in conversations.
\newblock \emph{arXiv preprint arXiv:2310.11374}.

\bibitem[{Zhao et~al.(2022{\natexlab{a}})Zhao, Zhang, Hu, Liu, Jin, Wang, and Li}]{zhao2022m3ed}
Jinming Zhao, Tenggan Zhang, Jingwen Hu, Yuchen Liu, Qin Jin, Xinchao Wang, and Haizhou Li. 2022{\natexlab{a}}.
\newblock M3ed: Multi-modal multi-scene multi-label emotional dialogue database.
\newblock \emph{arXiv preprint arXiv:2205.10237}.

\bibitem[{Zhao et~al.(2022{\natexlab{b}})Zhao, Zhang, Hu, Liu, Jin, Wang, and Li}]{zhao-etal-2022-m3ed}
Jinming Zhao, Tenggan Zhang, Jingwen Hu, Yuchen Liu, Qin Jin, Xinchao Wang, and Haizhou Li. 2022{\natexlab{b}}.
\newblock \href {https://doi.org/10.18653/v1/2022.acl-long.391} {{M}3{ED}: Multi-modal multi-scene multi-label emotional dialogue database}.
\newblock In \emph{Proceedings of the 60th Annual Meeting of the Association for Computational Linguistics (Volume 1: Long Papers)}, pages 5699--5710, Dublin, Ireland. Association for Computational Linguistics.

\bibitem[{Zhong et~al.(2019)Zhong, Wang, and Miao}]{zhong2019knowledge}
Peixiang Zhong, Di~Wang, and Chunyan Miao. 2019.
\newblock Knowledge-enriched transformer for emotion detection in textual conversations.
\newblock \emph{arXiv preprint arXiv:1909.10681}.

\end{thebibliography}

\appendix



\end{document}